\documentclass[runningheads]{llncs}
\usepackage[T1]{fontenc}
\usepackage{graphicx}
\usepackage{xcolor}
\usepackage{amsmath}
\usepackage{amssymb}
\usepackage{array}
\usepackage{multirow}
\usepackage{color}
\usepackage{colortbl}
\usepackage{framed}
\usepackage{bm}
\usepackage{bbm}
\usepackage{xspace}
\usepackage{enumitem}
\usepackage{algorithm}
\usepackage{listings}
\usepackage{url}
\usepackage{booktabs}
\usepackage{pifont} 
\usepackage{lipsum}
\usepackage{bm}
\usepackage{orcidlink}
\usepackage{multirow}
\usepackage[table]{xcolor}
\usepackage{pifont}

\usepackage{amsmath}
\usepackage{amssymb}
\usepackage{algorithm}
\usepackage{algorithmicx}
\usepackage{algpseudocode}
\usepackage{booktabs}
\usepackage{hyperref} 

\newcommand{\tit}[1]{\smallbreak\noindent\textbf{#1.}}
\newcommand{\tinytit}[1]{\noindent\textbf{#1.}}

\def \ie {\emph{i.e.}}
\def \eg {\emph{e.g.}}

\definecolor{OurColor}{HTML}{E8EBFF}
\definecolor{TitleColor}{rgb}{0.95, 0.945, 0.975}
\definecolor{blond}{RGB}{255, 244, 214}  

\newcommand{\ours}{{GramSR}\xspace}

\begin{document}
\sloppy

\title{GramSR: Visual Feature Conditioning for Diffusion-Based Super-Resolution}
\titlerunning{GramSR: Visual Feature Conditioning for Diffusion-Based Super-Resolution}
%
\author{Fabio D'Oronzio \and
Federico Putamorsi\orcidlink{0009-0009-0720-8051} \and
Leonardo Zini\orcidlink{0009-0003-9439-9867}\and\\
Marcella Cornia\orcidlink{0000-0001-9640-9385} \and
Lorenzo Baraldi\orcidlink{0000-0001-5125-4957}
}
\authorrunning{F. D'Oronzio et al.}
%
\institute{University of Modena and Reggio Emilia, Italy\\
\email{\{name.surname\}@unimore.it}
\vspace{-0.3cm}
}
\maketitle              
\begin{abstract}
Despite recent advances, single-image super-resolution (SR) remains challenging, especially in real-world scenarios with complex degradations. Diffusion-based SR methods, particularly those built on Stable Diffusion, leverage strong generative priors but commonly rely on text conditioning derived from semantic captioning. Such textual descriptions provide only high-level semantics and lack the spatially aligned visual information required for faithful restoration, leading to a representation gap between abstract semantics and spatially aligned visual details. To address this limitation, we propose GramSR, a one-step diffusion-based SR framework that replaces text conditioning with dense visual features extracted from the low-resolution input using a pre-trained DINOv3 encoder. GramSR adopts a three-stage LoRA architecture, where pixel-level, semantic-level, and texture-level LoRA modules are trained sequentially. The pixel-level module focuses on degradation removal using $\ell_2$ loss, the semantic-level module enhances perceptual details via LPIPS and CSD losses, and the texture-level module enforces feature correlation consistency through a Gram matrix loss computed from DINOv3 features. At inference, independent guidance scales enable flexible control over degradation removal, semantic enhancement, and texture preservation. Extensive experiments on standard SR benchmarks demonstrate that GramSR consistently outperforms existing one-step diffusion-based methods, achieving superior structural fidelity and texture realism. The code for this work is available at: \url{https://github.com/aimagelab/GramSR}.

\vspace{-.1cm}
\end{abstract}

\section{Introduction}
\label{sec:intro}
\vspace{-.1cm}
Single-image super-resolution (SR) aims to recover high-resolution images from their low-resolution counterparts, remaining a fundamental problem in computer vision. Since SRCNN~\cite{dong2014learning}, deep learning-based SR methods have primarily optimized pixel-level fidelity metrics such as PSNR and SSIM~\cite{wang2004image} under simplified degradation assumptions. Subsequent works explored deeper and more expressive architectures~\cite{lim2017enhanced,tong2017image}, but these approaches often fail to generalize to real-world scenarios involving complex degradations. This limitation has motivated research on realistic degradation modeling~\cite{cai2019toward,wei2020component}, perceptual losses~\cite{johnson2016perceptual}, and GAN-based training~\cite{ledig2017photo,wang2021real,zhang2021designing}. Despite improved visual realism, adversarial methods are prone to training instability and artifact generation~\cite{liang2022details,xie2023desra}.

Diffusion models~\cite{dhariwal2021diffusion,song2020score} have emerged as a powerful alternative, offering expressive generative priors and stable optimization. Large-scale text-to-image models, particularly Stable Diffusion~\cite{rombach2022high}, demonstrate remarkable ability in generating semantically rich details, motivating their use in SR~\cite{wang2024exploiting,wu2024seesr,yang2024pixel,yu2024scaling}. Further efforts focus on improving efficiency through paired data training~\cite{yue2023resshift}, degradation-aware modeling~\cite{Li2024BlindDiffED}, flexible control mechanisms~\cite{chai2025omniscales,Wang2024SAMDiffSRSD}, and quality-aware optimization~\cite{Wu2025DP2OSRDP}. However, multi-step diffusion sampling incurs substantial computational cost, motivating one-step approaches~\cite{arora2025guidesrrethinkingguidanceonestep,Chen2024AdversarialDC,Sun2025PocketSRTS,sun2025pixel,wang2024sinsr,wu2024one} that achieve real-time performance via distillation and architectural simplification.

A key limitation of existing diffusion-based SR methods lies in their reliance on \emph{text conditioning}~\cite{sun2025pixel,wu2024one,wu2024seesr}. While effective for text-to-image generation, textual descriptions obtained from captioning models~\cite{li2023blip2,li2022blip} or tagging modules~\cite{wu2024seesr} encode only high-level semantic concepts, such as object categories and scene context. They lack spatially aligned structural cues, fine-grained geometry, and local texture patterns that are critical for faithful image restoration. This modality mismatch introduces a semantic gap between abstract linguistic representations and the precise visual guidance required for SR. Moreover, current training objectives mainly emphasize pixel-level fidelity and semantic consistency~\cite{sun2025pixel}, without explicitly modeling texture statistics characterized by repeated patterns and feature correlations.

To address these challenges, we propose \textbf{\ours}, a one-step diffusion-based SR framework with two key innovations. First, we replace text conditioning with dense visual features extracted directly from the low-resolution input using a pre-trained DINOv3 encoder~\cite{simeoni2025dinov3}. This provides spatially aligned, hierarchical visual representations that capture both local details and global context. Second, we extend prior dual-LoRA frameworks~\cite{sun2025pixel,wu2024one} with a third, texture-level LoRA module optimized via Gram matrix loss to preserve second-order feature statistics. While the pixel-level and semantic-level LoRAs address degradation removal and perceptual detail generation, respectively, the texture-level LoRA enforces feature correlation consistency by aligning Gram matrices between the super-resolved output and the ground truth. This sequential training strategy disentangles complementary restoration objectives, while inference-time guidance scales enable independent control over degradation removal, semantic enhancement, and texture preservation. Extensive experiments on standard SR benchmarks demonstrate that \ours consistently outperforms existing one-step diffusion-based methods in both quantitative metrics and perceptual quality, achieving notably improved texture fidelity and structural consistency.

\section{Related Work}
\label{sec:related}

\tinytit{Early Super-Resolution Methods}
Since the introduction of SRCNN~\cite{dong2014learning}, deep learning-based methods have become the dominant paradigm for SR. Early approaches mainly optimized pixel-level fidelity metrics~\cite{wang2004image} under simplified and known degradations (\eg, bicubic downsampling). Later works explored more advanced architectures with improved connectivity, hierarchical representations, and global context modeling~\cite{tong2017image,lim2017enhanced}. Despite strong performance on synthetic benchmarks, these methods often struggle to generalize to real-world low-quality (LQ) images with complex and unknown degradations. To address this limitation, real-world SR has been studied through the collection of paired LQ-HQ datasets~\cite{cai2019toward,wei2020component} or the synthesis of more realistic training data via degradation modeling. With the rise of generative approaches, perceptual losses~\cite{johnson2016perceptual} and GAN-based training~\cite{ledig2017photo} have been widely adopted to enhance visual realism. Representative methods such as BSRGAN~\cite{zhang2021designing} and Real-ESRGAN~\cite{wang2021real} employ complex, randomized degradation pipelines to generate realistic LQ-HQ pairs. However, adversarial training is often unstable and prone to artifacts, motivating subsequent efforts to improve robustness and visual quality~\cite{liang2022details,xie2023desra}.

\tit{Diffusion-based Super-Resolution Methods}
Diffusion models (DMs) have recently gained attention for SR due to their strong generative priors and stable training dynamics. Early DM-based methods adapt denoising diffusion probabilistic models (DDPMs)~\cite{dhariwal2021diffusion,song2020score} via gradient guidance, enabling training-free restoration under simple degradations, but showing limited robustness to real-world degradations. More recent approaches leverage large-scale text-to-image models, particularly Stable Diffusion~\cite{rombach2022high}, as powerful semantic priors. Representative methods include StableSR~\cite{wang2024exploiting}, PASD~\cite{yang2024pixel}, and SeeSR~\cite{wu2024seesr}, which introduce different conditioning and feature-integration strategies to improve robustness and detail generation. Other works explore truncating the diffusion process or enhancing semantic consistency, such as CCSR~\cite{sun2023improving} and SUPIR~\cite{yu2024scaling}.

A major drawback of diffusion-based SR is the high inference cost caused by multi-step sampling. To improve efficiency, recent methods investigate alternative training and inference strategies, including paired-data training~\cite{yue2023resshift}, joint degradation estimation~\cite{Li2024BlindDiffED}, and flexible control mechanisms~\cite{chai2025omniscales,Wang2024SAMDiffSRSD}. Additional efforts incorporate perceptual quality optimization into diffusion training~\cite{Wu2025DP2OSRDP}. Nevertheless, most existing approaches still rely on text conditioning and multi-step sampling, motivating one-step diffusion frameworks with more effective and visually grounded conditioning mechanisms.

\tit{One-Step Super-Resolution Methods} 
Motivated by the high computational cost of multi-step diffusion sampling, recent works have explored one-step diffusion-based SR to achieve efficient restoration with a single forward pass. SinSR~\cite{wang2024sinsr} performs one-step distillation from a multi-step teacher but often yields over-smoothed results due to limited detail recovery. OSEDiff~\cite{wu2024one} improves efficiency by directly conditioning on LQ images and distilling diffusion trajectories. GuideSR~\cite{arora2025guidesrrethinkingguidanceonestep} introduces dual-branch guidance to enhance structural fidelity, while AdcSR~\cite{Chen2024AdversarialDC} combines diffusion and adversarial training for compact inference. PocketSR~\cite{Sun2025PocketSRTS} further targets lightweight deployment with simplified architectures. More recently, PiSA-SR~\cite{sun2025pixel} proposes a dual-LoRA framework to balance pixel-level fidelity and semantic enhancement with controllable guidance.

Despite these advances, existing one-step methods largely rely on text-based conditioning and limited loss formulations, which restrict their ability to preserve spatially aligned structures and fine-grained textures under complex real-world degradations. In contrast, our method replaces text conditioning with dense visual features and explicitly models texture consistency through Gram-based alignment, enabling more faithful structure and texture restoration within an efficient one-step diffusion framework.

\section{\ours}
\label{sec:method}

This section presents \ours, a one-step diffusion-based SR framework that replaces text conditioning with visual feature conditioning and introduces texture-aware adaptation via Gram matrix alignment. We first describe the residual formulation of one-step diffusion SR, then detail the proposed visual conditioning mechanism, the three-stage LoRA training strategy, and finally the inference procedure with triple guidance control.

\subsection{Preliminaries}

We formulate the SR problem as residual learning in the latent space of a pre-trained VAE. Let $x_L$ and $x_H$ denote the low-quality (LQ) and high-quality (HQ) images, respectively, and let $\mathcal{E}$ and $\mathcal{D}$ be the frozen VAE encoder and decoder. Their corresponding latent representations are $z_L = \mathcal{E}(x_L)$ and $z_H = \mathcal{E}(x_H)$. Following the one-step diffusion paradigm, SR is achieved by directly transforming the LQ latent into the HQ latent through a single denoising step:
\begin{equation}
z_H = z_L - \epsilon_\theta(z_L, c),
\end{equation}
where $\epsilon_\theta$ is a diffusion denoising network parameterized by $\theta$, and $c$ denotes conditioning information. Unlike multi-step diffusion models that iteratively refine Gaussian noise, this formulation directly learns the residual between $z_L$ and $z_H$, allowing efficient and stable training while focusing the model capacity on recovering high-frequency details.

Previous approaches typically derive $c$ from text prompts generated by image captioning or tagging models. Such textual conditioning provides high-level semantic cues but lacks spatial alignment with the input image, limiting its ability to preserve fine-grained structures. To overcome this limitation, we replace text conditioning with dense visual features extracted directly from the LQ image.

\begin{figure}[t]
\centering
\includegraphics[width=\linewidth]{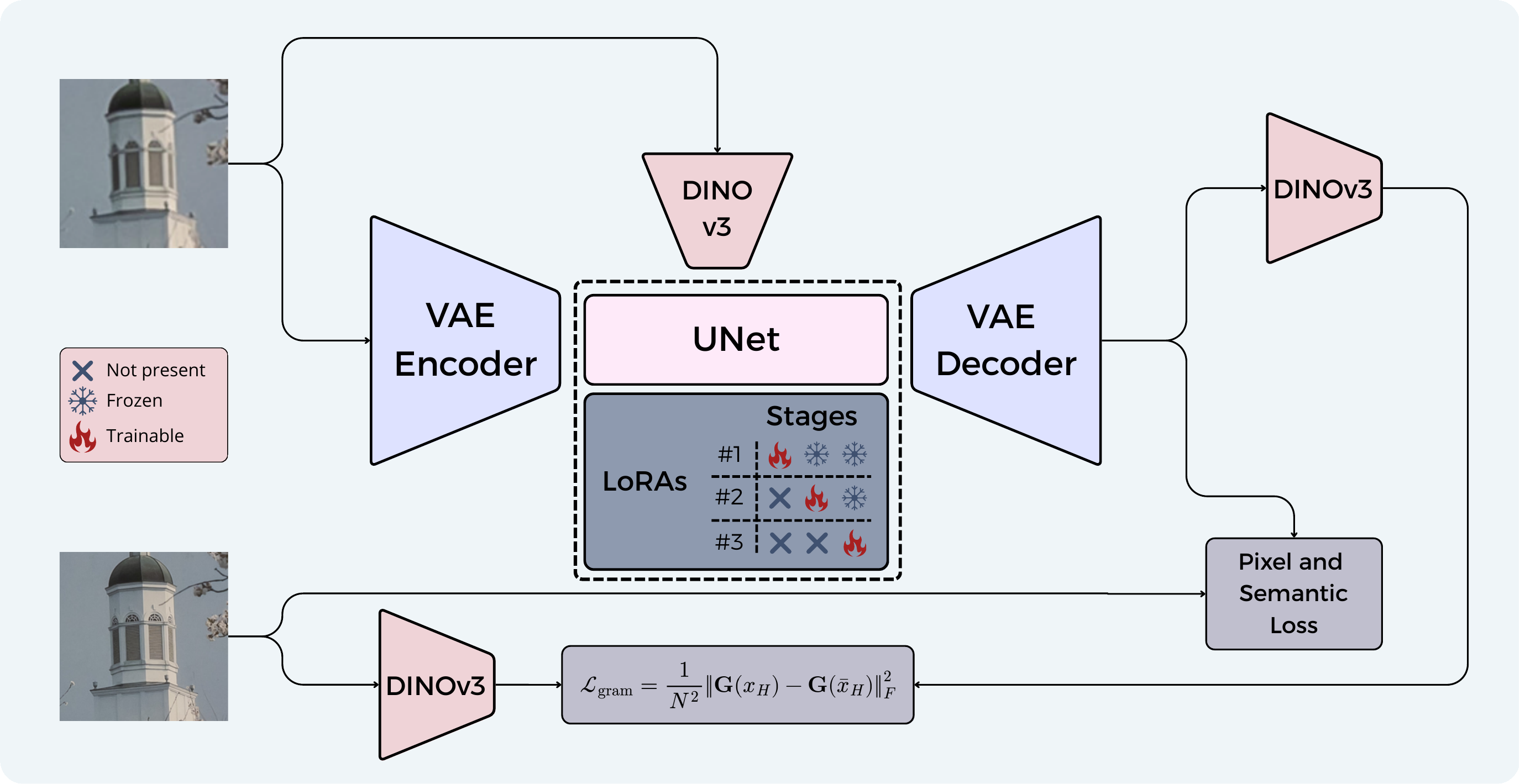}
\vspace{-0.55cm}
\caption{Overview of the proposed three-stage training framework. The architecture consists of a frozen DINOv3~\cite{simeoni2025dinov3} encoder for visual conditioning, a frozen VAE encoder-decoder pair, and a diffusion U-Net equipped with three sequential LoRA~\cite{hu2022lora} modules. In Stage~1, the pixel-level LoRA is trained with pixel-wise loss. In Stage~2, the pixel-level LoRA is frozen and the semantic-level LoRA is trained with perceptual and semantic losses. In Stage~3, both are frozen, and the texture-level LoRA is trained with Gram matrix loss to enhance texture consistency. The DINOv3 encoder remains frozen throughout all stages, and all LoRA modules are applied only to the U-Net module.}
\label{fig:architecture}
\vspace{-0.35cm}
\end{figure}

\subsection{Visual Conditioning}
We employ a pre-trained DINOv3 encoder~\cite{simeoni2025dinov3} to extract patch-level visual features from the input $x_L$:
\begin{equation}
\mathbf{F}_{\text{ViT}} = \text{DINOv3}(x_L) \in \mathbb{R}^{N \times d},
\end{equation}
where $N$ is the number of patches and $d$ is the feature dimension. These features encode spatially aligned visual information ranging from local textures to global context. The DINOv3 encoder is kept frozen throughout training.

To integrate the extracted features into the diffusion model, we introduce a lightweight MLP adapter that projects $\mathbf{F}_{\text{ViT}}$ into the conditioning space of the diffusion U-Net. Specifically, the adapter consists of two linear transformations with a ReLU activation in between. 
The adapter is trained jointly with the LoRA modules, which are attached to the convolutional and MLP layers of the denoising U-Net, in the first two stages, allowing the model to learn an effective projection of visual features for SR conditioning.

Compared to text conditioning, visual conditioning preserves spatial correspondence and avoids the bottleneck of intermediate captioning, providing more precise guidance for structure and texture restoration.

\subsection{Three-Stage LoRA Training}

To disentangle different restoration objectives, we adopt a sequential LoRA-based training strategy that decomposes SR into pixel-level, semantic-level, and texture-level enhancement. All LoRA modules are applied to the U-Net of the diffusion model (\ie, Stable Diffusion in our experiments), while the backbone parameters remain frozen.

\tit{Overall Architecture}
The model consists of three LoRA modules: a pixel-level LoRA $\Delta\theta_{\text{pix}}$, a semantic-level LoRA $\Delta\theta_{\text{sem}}$, and a texture-level LoRA $\Delta\theta_{\text{gram}}$. Each module is optimized in a dedicated training stage, with previously trained modules frozen to avoid interference across objectives.

\tit{Pixel-Level LoRA Module}
In the first stage, we train $\Delta\theta_{\text{pix}}$ to remove degradations such as noise, blur, and downsampling artifacts. The optimization uses an $\ell_2$ loss between the reconstructed image and the ground truth (\ie, $\mathcal{L}_{\text{MSE}}$), encouraging accurate pixel-level reconstruction and stable convergence.

\tit{Semantic-Level LoRA Module}
In the second stage, $\Delta\theta_{\text{pix}}$ is frozen and a semantic-level LoRA $\Delta\theta_{\text{sem}}$ is introduced. This module enhances perceptual realism by optimizing a combination of LPIPS~\cite{zhang2018unreasonable} and classifier score distillation (CSD)~\cite{yu2024texttod} losses (\ie, $\mathcal{L}_{\text{LPIPS}}$ and $\mathcal{L}_{\text{CSD}}$, respectively). These objectives encourage the generation of semantically rich and visually plausible details while preserving the pixel-level corrections learned in the first stage. Following previous works~\cite{sun2025pixel}, $\mathcal{L}_{\text{LPIPS}}$ and $\mathcal{L}_{\text{CSD}}$ are jointly optimized with $\mathcal{L}_{\text{MSE}}$ during this stage to maintain pixel-level fidelity.

\tit{Texture-level LoRA Module}
While the semantic-level LoRA generates perceptually realistic details, it does not explicitly enforce texture consistency between the SR output and ground truth. In particular, textures in natural images exhibit statistical regularities that can be captured by second-order statistics. The Gram matrix, originally popularized in neural style transfer~\cite{gatys2016image} and more recently used in~\cite{simeoni2025dinov3}, provides an effective representation of texture by computing correlations between spatial locations.

Given DINOv3 features $\mathbf{F} \in \mathbb{R}^{N \times d}$ extracted from an image, we first normalize them and compute the Gram matrix over spatial locations:
\begin{equation}
\hat{\mathbf{F}} = \frac{\mathbf{F}}{\|\mathbf{F}\|_2}, \quad
G_{ij} = \sum_{k=1}^{d} \hat{F}_{i,k} \cdot \hat{F}_{j,k},
\end{equation}
where $G_{ij}$ measures the correlation between the $i$-th and $j$-th patches. The resulting Gram matrix encodes global texture statistics that are invariant to local spatial arrangements.

During the third training stage, both $\Delta\theta_{\text{pix}}$ and $\Delta\theta_{\text{sem}}$ are frozen, and only $\Delta\theta_{\text{gram}}$ is optimized by minimizing the discrepancy between the Gram matrices of the super-resolved output and the ground-truth image. Specifically, we extract DINOv3 features from both the SR output $x_H$ and the ground-truth $\bar{x}_H$, compute their respective Gram matrices, and minimize their difference. Formally, this is defined as:
\begin{equation}
\mathcal{L}_{\text{gram}} =
\frac{1}{N^2} \left\| \mathbf{G}(x_H) -
\mathbf{G}(\bar{x}_H) \right\|^2_F.
\end{equation}
where $\left\|\cdot\right\|_F$ denotes the Frobenius norm.

The texture-level LoRA module is trained using a joint loss function:
\begin{equation}\label{eq:training_loss}
\mathcal{L} =
\lambda_1 \mathcal{L}_{\text{MSE}} +
\lambda_2 \mathcal{L}_{\text{LPIPS}} +
\lambda_3 \mathcal{L}_{\text{CSD}} +
\lambda_4 \mathcal{L}_{\text{gram}},
\end{equation}
where $\lambda_1$, $\lambda_2$, $\lambda_3$, and $\lambda_4$ balance the contributions of pixel fidelity, perceptual quality, semantic consistency, and texture alignment, ensuring that texture enhancement does not interfere with pixel-level fidelity and semantic content.

\subsection{Inference with Triple Guidance}
At inference time, the three LoRA modules enable flexible and interpretable control over the restoration process. The denoising prediction is computed as:
\begin{equation}
\label{eq:inference}
\epsilon_\theta(z_L) =
\epsilon_{\theta_0}(z_L) +
\lambda_{\text{pix}} \Delta\theta_{\text{pix}}(z_L) +
\lambda_{\text{sem}} \Delta\theta_{\text{sem}}(z_L) +
\lambda_{\text{gram}} \Delta\theta_{\text{gram}}(z_L),
\end{equation}
where $\epsilon_{\theta_0}$ is the output of the frozen backbone. The delta terms are defined as:
\begin{equation}
\Delta\theta_i(z_L) = \begin{cases}
\epsilon_{\theta_{\text{pix}}}(z_L), & i = \text{pix} \\
\epsilon_{\theta_i}(z_L) - \epsilon_{\theta_{i-1}}(z_L), & i \in \{\text{sem}, \text{gram}\}
\end{cases}
\end{equation}
in which $\theta_0$ is the base model, and the subscripts follow the order: $0 \to \text{pix} \to \text{sem} \to \text{gram}$ controlling the strengths of degradation removal, semantic enhancement, and texture preservation, respectively.

\section{Experiments}

\subsection{Experimental Settings}
\tinytit{Training and Evaluation Datasets}
To ensure consistency with existing diffusion-based SR methods, we follow an established data preparation and evaluation protocol. In particular, the training corpus is composed of images from LSDIR~\cite{li2023lsdir} together with the first 10k samples of FFHQ~\cite{bai2023ffhq}, offering a wide range of natural image content. LQ-HQ training pairs are generated synthetically by applying the degradation process introduced by Real-ESRGAN~\cite{wang2021real}, which emulates complex real-world degradations.

The evaluation is conducted on standard synthetic and real-world datasets, following the evaluation protocol defined in~\cite{wu2024one}. For synthetic testing, we employ 3,000 samples from DIV2K~\cite{agustsson2017ntire}, cropping images at a resolution of 512$\times$512 and degrading them using the Real-ESRGAN degradation pipeline~\cite{wang2021real}. Real-world evaluation relies on samples from the RealSR~\cite{cai2019toward} and DRealSR~\cite{wei2020component} datasets, where LQ images are center-cropped to 128$\times$128 and their corresponding HQ counterparts to 512$\times$512. 

\tit{Evaluation Metrics}
To ensure an accurate evaluation of the quality of the image generation, we employ standard metrics, typically used to measure the performance of SR methods. Specifically, image reconstruction fidelity is measured using PSNR and SSIM~\cite{wang2004image}, computed on the luminance channel in the YCbCr color space. Perceptual quality is assessed in the RGB space through LPIPS~\cite{zhang2018unreasonable} and DISTS~\cite{ding2020image}. In addition, FID~\cite{heusel2017gans} is reported to quantify the distribution gap between restored images and their ground-truth counterparts. For no-reference assessment, NIQE~\cite{mittal2012making} is employed to evaluate the perceptual naturalness of the super-resolved outputs.

\tit{Training Details}
The proposed method is trained for the $\times4$ SR setting on a pre-trained Stable Diffusion v2.1~\cite{rombach2022high}. For visual conditioning, we use the DINOv3~\cite{simeoni2025dinov3} ViT-B with feature dimensionality equal to 768, while for Gram matrix alignment, we use the DINOv3 ViT-S+ with feature dimensionality of 384. To enable efficient adaptation, lightweight LoRA modules~\cite{hu2022lora} with rank $4$ are inserted into the convolutional and MLP layers of the denoising U-Net.
During training, input images are randomly cropped into patches of size 512$\times$512.

The training process is divided into two steps. First, we train the pixel-level LoRA $\Delta\theta_\text{pix}$ and semantic-level LoRA $\Delta\theta_\text{sem}$ with a learning rate of $5 \times 10^{-5}$ employing the same number of iterations as proposed in~\cite{sun2025pixel}. Then, we train the LoRA $\Delta\theta_\text{gram}$ with a learning rate of $5 \times 10^{-6}$, using the NIQE score~\cite{mittal2012making} on the validation set as an early stopping condition. Training is performed using Adam as optimizer with a batch size of 16 on a single NVIDIA L40S GPU. Where not stated otherwise, the loss weights in Eq.~\ref{eq:training_loss} are set to $\lambda_1 = \lambda_3 = 1.0$, $\lambda_2 = 2.0$, and $\lambda_4 = 500.0$, empirically chosen to balance the magnitudes of different loss terms. Instead, the weights of Eq.~\ref{eq:inference} that serve to balance the contribution of each LoRA module are set to 1. 

\begin{table}[t]
\caption{Comparison of one-step diffusion-based SR methods across different datasets.}
\vspace{-.6em}
\centering
\setlength{\tabcolsep}{.55em}
\renewcommand{\arraystretch}{0.95}
\resizebox{0.9\linewidth}{!}{
\begin{tabular}{lcccccc}
\toprule
&\multicolumn{6}{c}{\textbf{DIV2K}} \\
\cmidrule{2-7}
 & PSNR $\uparrow$ & SSIM $\uparrow$ & LPIPS $\downarrow$ & DISTS $\downarrow$ & FID $\downarrow$ & NIQE $\downarrow$ \\
 \midrule
AdcSR~\cite{Chen2024AdversarialDC} & 23.74 & 60.17 & 28.53 & 18.99 & 25.52 & \textbf{4.36}\\
S3Diff~\cite{zhang2024degradation} & 23.53 & 59.33 & 25.81 & \textbf{17.30} & \textbf{19.51} & 4.74\\ 
SinSR~\cite{wang2024sinsr} & \underline{24.29} & 60.08 & 32.28 & 20.49 & 37.17 & 5.79\\
OSEDiff~\cite{wu2024one} & 23.72 & \underline{61.09} & 29.42 & 19.75 & 26.34 & 4.71\\
PiSA-SR~\cite{sun2025pixel} & 23.86 &  60.57 & \underline{28.21} & 19.30 & 25.07 & \underline{4.57}\\
\rowcolor{OurColor} \textbf{\ours (Ours)} & \textbf{24.79} & \textbf{63.31} & \textbf{27.18} & \underline{18.38} & \underline{21.84} & 4.60\\
\midrule
&\multicolumn{6}{c}{\textbf{RealSR}} \\
\cmidrule{2-7}
 & PSNR $\uparrow$ & SSIM $\uparrow$ & LPIPS $\downarrow$ & DISTS $\downarrow$ & FID $\downarrow$ & NIQE $\downarrow$ \\
\midrule
AdcSR~\cite{Chen2024AdversarialDC} & 25.47 & 73.01 & 28.85 & 21.29 & 118.42 & \underline{5.35}\\
S3Diff~\cite{zhang2024degradation} & 25.18 & 72.69 & 27.21 & \underline{20.05} & \underline{105.09} & \textbf{5.27}\\ 
SinSR~\cite{wang2024sinsr} & \underline{26.18} & 73.85 & 30.50 & 23.16 & 136.16 & 6.03\\
OSEDiff~\cite{wu2024one} & 25.15 & 73.41 & 29.20 & 21.27 & 123.50 & 5.64\\
PiSA-SR~\cite{sun2025pixel} & 25.48 & \underline{74.15} & \underline{26.72} & 20.42 & 124.34 & 5.51\\
\rowcolor{OurColor} \textbf{\ours (Ours)} & \textbf{26.81} & \textbf{76.68} & \textbf{23.05} & \textbf{17.91} & \textbf{100.71} & 5.84 \\
\midrule
&\multicolumn{6}{c}{\textbf{DRealSR}} \\
\cmidrule{2-7}
 & PSNR $\uparrow$ & SSIM $\uparrow$ & LPIPS $\downarrow$ & DISTS $\downarrow$ & FID $\downarrow$ & NIQE $\downarrow$ \\
 \midrule
AdcSR~\cite{Chen2024AdversarialDC} & 28.10 & 77.26 & 30.46 & 22.00 & 134.01 & 6.45 \\
S3Diff~\cite{zhang2024degradation} & 27.54 & 74.90 & 31.09 & \underline{21.00} & \underline{118.59} & \underline{6.21}\\ 
SinSR~\cite{wang2024sinsr} & 28.22 & 75.53 & 35.02 & 24.77 & 176.84 & 6.71\\
OSEDiff~\cite{wu2024one} & 27.92 & \underline{78.35} & 29.67 & 21.62 & 135.53 & 6.44\\
PiSA-SR~\cite{sun2025pixel} & \underline{28.23} & 77.97 & \underline{29.63} & 21.69 & 130.61 & \textbf{6.20}\\
\rowcolor{OurColor} \textbf{\ours (Ours)} & \textbf{29.69} & \textbf{81.21} & \textbf{26.03} & \textbf{19.12} & \textbf{117.85} & 6.42\\
\bottomrule
\end{tabular}
}
\label{tab:comparison}
\vspace{-.4cm}
\end{table}

\subsection{Comparison with the State of the Art}

\tinytit{Quantitative Evaluation}
We compare our method with recent state-of-the-art one-step diffusion-based SR approaches, including S3Diff~\cite{zhang2024degradation}, SinSR~\cite{wang2024sinsr}, OSEDiff~\cite{wu2024one}, PiSA-SR~\cite{sun2025pixel}, and AdcSR~\cite{Chen2024AdversarialDC}. 
All methods follow a similar diffusion formulation and are evaluated under the same testing protocol as our proposed \ours. Results are reported in Table~\ref{tab:comparison}, comparing \ours and competitors on the DIV2K~\cite{agustsson2017ntire}, RealSR~\cite{cai2019toward}, and DRealSR~\cite{wei2020component} datasets.

Overall, \ours achieves the best performance across most metrics and datasets, demonstrating consistent improvements in both reconstruction fidelity and perceptual quality. On DIV2K, our method reaches the highest PSNR and SSIM while also reducing LPIPS compared to prior approaches, indicating more accurate detail recovery without compromising visual realism. On real-world benchmarks, the advantages are even more pronounced: \ours yields clear gains on RealSR and DRealSR, achieving the best PSNR and SSIM and substantially lower LPIPS and DISTS, which reflect improved texture coherence and perceptual consistency under complex degradations.

In particular, compared to PiSA-SR, the most closely related method that employs dual-LoRA adaptation with text-based conditioning, \ours consistently delivers superior results across all datasets. The improvements are especially notable on RealSR and DRealSR, where \ours reduces LPIPS, DISTS, and FID by a significant margin while also improving PSNR and SSIM. This indicates that the proposed contributions lead to more faithful reconstruction of fine-grained structures and textures, which are insufficiently captured by the two-stage, text-conditioned baseline.
Overall, while some competing methods achieve competitive scores on individual perceptual metrics, \ours provides a more favorable and consistent trade-off between fidelity- and perception-oriented measures. These results demonstrate the robustness of our approach and its effectiveness in handling real-world SR scenarios.

\begin{figure}[t]
    \centering
    \includegraphics[width=\linewidth]{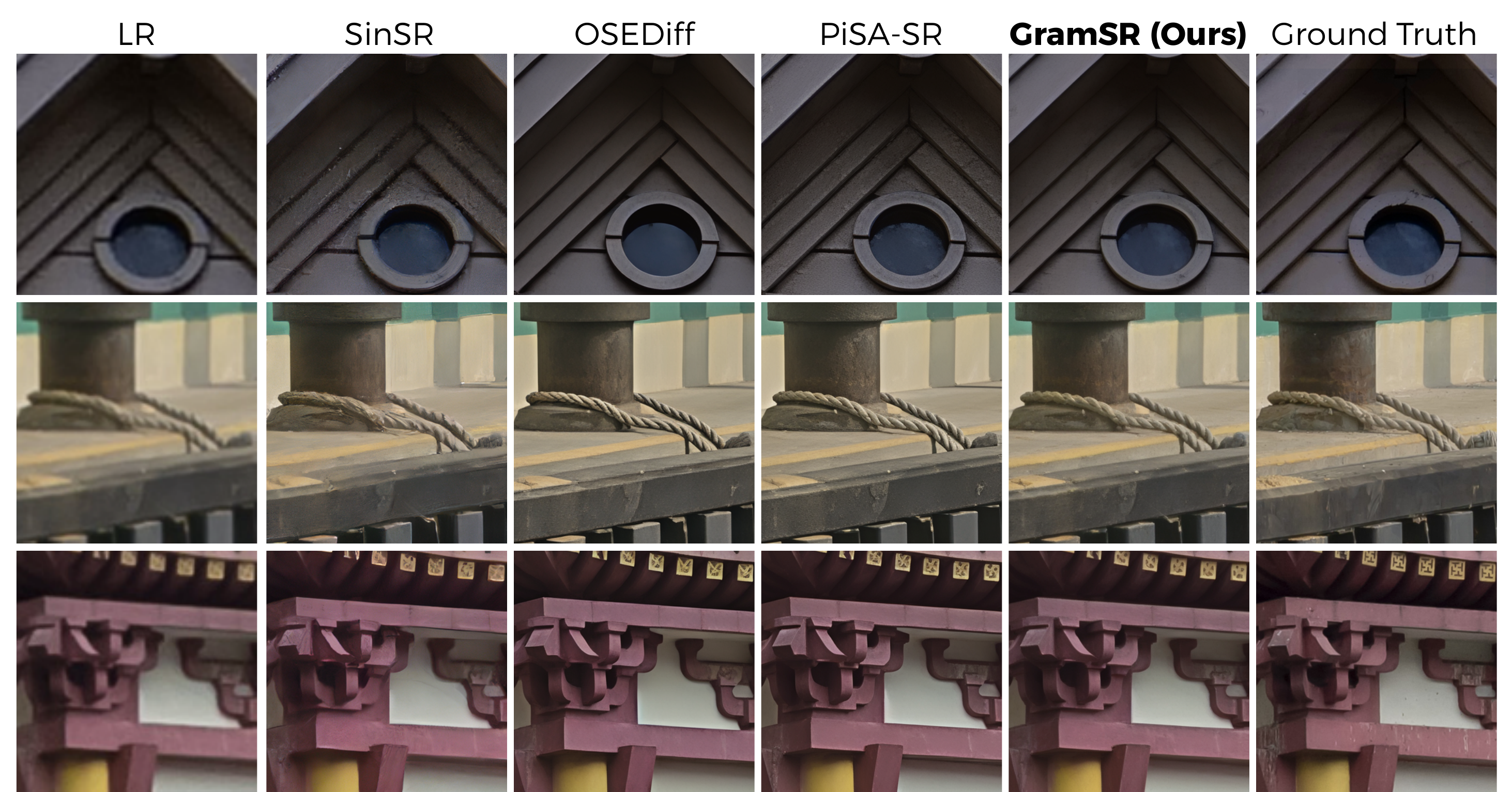}
    \vspace{-0.6cm}
    \caption{Qualitative comparison on real-world images from the RealSR dataset. From left to right: low-resolution input, results of SinSR~\cite{wang2024sinsr}, OSEDiff~\cite{wu2024one}, PiSA-SR~\cite{sun2025pixel}, \ours (Ours), and ground truth. 
    }
    \label{fig:qualitatives}
    \vspace{-0.6cm}
\end{figure}

\tit{Qualitative Evaluation} Fig.~\ref{fig:qualitatives} presents a qualitative comparison on real-world images from the RealSR dataset. The first column shows the low-resolution input, followed by the results of different one-step diffusion-based methods, with the ground truth shown in the last column. As it can be seen, OSEDiff~\cite{wu2024one} tends to produce over-smoothed reconstructions, resulting in the loss of fine structural details. SinSR~\cite{wang2024sinsr} generates sharper outputs but often exhibits inconsistent textures and local artifacts in regions with complex patterns. PiSA-SR~\cite{sun2025pixel} improves semantic detail generation; however, some repetitive textures and fine structures remain imperfectly reconstructed. In contrast, \ours produces visually more coherent results, with improved preservation of both structural details and texture patterns. The advantages are particularly evident in challenging regions, where our method yields more consistent textures and fewer artifacts, resulting in outputs that more closely resemble the ground truth.

\subsection{Ablation Studies}

\begin{table}[t]
\centering
\caption{Ablation study results on the effects of visual conditioning and texture-level LoRA with Gram matrix loss, reported on the DIV2K, RealSR, and DRealSR datasets.}
\vspace{-.6em}
\setlength{\tabcolsep}{.35em}
\resizebox{\linewidth}{!}{
\begin{tabular}{lllcccccc}
\toprule
& & & PSNR $\uparrow$ & SSIM $\uparrow$ & LPIPS $\downarrow$ & DISTS $\downarrow$ & FID $\downarrow$ & NIQE $\downarrow$ \\
 \midrule
& & Baseline  & 23.86 &  60.57 & 28.21 & 19.30 & 25.07 & \textbf{4.57}\\
& & \hspace{0.3cm}$+$ Visual Conditioning  & 24.62 & 63.05 & 27.90 & 19.03 & 24.15 & 4.79\\ 
& & \hspace{0.3cm}$+\Delta\theta_{gram}$ (Third LoRA) & \underline{24.76} & \underline{63.38} & \underline{27.10} & 1\underline{8.50} & \underline{22.56} & 4.68\\ 
\rowcolor{OurColor}
\cellcolor{white}\multirow{-4}{*}{{\rotatebox[origin=c]{90}{\textbf{DIV2K}}}} & \cellcolor{white} & \textbf{\ours (Ours)} & \textbf{24.79} & \textbf{63.31} & \textbf{27.18} & \textbf{18.38} & \textbf{21.84} & \underline{4.60}\\
\midrule
& & Baseline & 25.48 & 74.15 & 26.72 & 20.42 & 124.34 & \textbf{5.51}\\
& & \hspace{0.3cm}$+$ Visual Conditioning  & \underline{26.59} & 76.01 & \underline{23.70} & \underline{18.35} & \underline{107.08} & 5.71\\ 
& & \hspace{0.3cm}$+\Delta\theta_{gram}$ (Third LoRA)  & \underline{26.59} & \underline{76.15} & 24.86 & 19.05 & 109.32 & \underline{5.69}\\ 
\rowcolor{OurColor}
\cellcolor{white}\multirow{-4}{*}{{\rotatebox[origin=c]{90}{\textbf{RealSR}}}} & \cellcolor{white} & \textbf{\ours (Ours)} & \textbf{26.81} & \textbf{76.68} & \textbf{23.05} & \textbf{17.91} & \textbf{100.71} & 5.84 \\
\midrule
& & Baseline & 28.23 & 77.97 & 29.63 & 21.69 & 130.61 & \textbf{6.20}\\
& & \hspace{0.3cm}$+$ Visual Conditioning  & \underline{29.56} & \underline{81.04} & \underline{26.04} & \underline{19.56} & 126.22 & 6.55\\ 
& & \hspace{0.3cm}$+\Delta\theta_{gram}$ (Third LoRA) & 29.44 &  80.61 & 27.37 & 20.21 & \underline{123.23} & \underline{6.42} \\
\rowcolor{OurColor}
\cellcolor{white}\multirow{-4}{*}{{\rotatebox[origin=c]{90}{\textbf{DRealSR}}}} & \cellcolor{white} & \textbf{\ours (Ours)}  & \textbf{29.69} & \textbf{81.21} & \textbf{26.03} & \textbf{19.12} & \textbf{117.85} & \underline{6.42}\\
\bottomrule
\end{tabular}
}
\label{tab:model_ablation}
\vspace{-0.35cm}
\end{table}

%
%

\tinytit{Effect of Individual Design Choices}
Table~\ref{tab:model_ablation} validates the contributions of the two key components of \ours (\ie, visual conditioning and the texture-level LoRA optimized with Gram matrix loss). The baseline corresponds to the dual-LoRA framework without visual conditioning and without the texture-level module. As shown, replacing text conditioning with visual conditioning consistently improves performance over the baseline, yielding higher PSNR and SSIM and lower perceptual errors on most settings. This demonstrates that dense visual features provide more effective and spatially aligned guidance for the task. Adding the texture-level LoRA further improves perceptual quality, leading to notable reductions in LPIPS, DISTS, and FID across datasets, regardless of the conditioning strategy. Combining both components produces the best overall results, confirming their complementary roles in improving structural fidelity and texture realism, particularly on real-world data.

\begin{table}[t]
\centering
\caption{Effect of the texture-level LoRA during training and inference on the RealSR dataset. Top: varying the Gram matrix loss weight $\lambda_4$ in Eq.~\ref{eq:training_loss}. Bottom: varying the texture-level guidance scale $\lambda_{\text{gram}}$ in Eq.~\ref{eq:inference}.}
\vspace{-.6em}
\setlength{\tabcolsep}{0.6em}
\resizebox{0.85\linewidth}{!}{
\begin{tabular}{l@{\hspace{0.2em}}lccccccc}
\toprule
& & & PSNR $\uparrow$ & SSIM $\uparrow$ & LPIPS $\downarrow$ & DISTS $\downarrow$ & FID $\downarrow$ & NIQE $\downarrow$ \\
\midrule
& & 50  & 26.59 & 75.98 & 23.54 & 18.20 & 103.09 & \textbf{5.63}  \\
& & 125 & 26.65 & 76.08 & 23.43 & \underline{18.09} & \underline{101.60} & \underline{5.65}  \\
& & 250 & \underline{26.69} & \underline{76.28} & \underline{23.48} & 18.15 & 102.15 & 5.77  \\
\multirow{-4}{*}{{\rotatebox[origin=c]{90}{\textbf{Training}}}} &\multirow{-4}{*}{{\rotatebox[origin=c]{90}{($\lambda_4$)}}} & 500 & \textbf{26.81} & \textbf{76.68} & \textbf{23.05} & \textbf{17.91} & \textbf{100.71} & 5.84 \\
\midrule
& & 0.25    & 26.72 & 76.35 & 23.54 & 18.24 & 105.65 & \textbf{5.78} \\
& & 0.50    & \underline{26.81} & 76.59 & 23.37 & 18.13 & 103.54 & \underline{5.84} \\
& & 0.75    & \textbf{26.84} & \textbf{76.69} & \underline{23.21} & \underline{18.00} & \underline{101.72} & 5.90 \\
\multirow{-4}{*}{{\rotatebox[origin=c]{90}{\textbf{Inference}}}} &\multirow{-4}{*}{{\rotatebox[origin=c]{90}{($\lambda_\text{gram}$)}}}  & 1.0     & \underline{26.81} & \underline{76.68} & \textbf{23.05} & \textbf{17.91} & \textbf{100.71} & \underline{5.84} \\
\bottomrule
\end{tabular}
}
\label{tab:lamda}
\vspace{-0.15cm}
\end{table}

\tit{Effect of Texture-Level Loss Weight and Guidance Scale} 
Table~\ref{tab:lamda} analyzes the influence of the texture-level LoRA at both training and inference. Specifically, we first study the effect of the Gram matrix loss weight $\lambda_4$ in the training objective of Eq.~\ref{eq:training_loss} (top rows of Table~\ref{tab:lamda}). Increasing $\lambda_4$ progressively improves both distortion- and perception-oriented metrics, with consistent reductions in LPIPS, DISTS, and FID as the weight increases. The best overall performance is achieved with $\lambda_4 = 500$, which yields the strongest perceptual improvements. These results indicate that emphasizing Gram-based feature correlation alignment during training effectively enhances texture consistency without degrading structural fidelity.

We further examine the role of the texture-level guidance scale $\lambda_{\text{gram}}$ in Eq.~\ref{eq:inference} during inference (bottom rows of Table~\ref{tab:lamda}), while keeping $\lambda_{\text{pix}}$ and $\lambda_{\text{sem}}$ fixed. Increasing $\lambda_{\text{gram}}$ strengthens texture enhancement, leading to improved perceptual metrics. In particular, $\lambda_{\text{gram}} = 1.0$ achieves the lowest LPIPS, DISTS, and FID, whereas a slightly smaller value of $0.75$ yields the best PSNR and SSIM. This behavior reveals a controllable trade-off between distortion fidelity and perceptual texture quality, allowing users to adjust the strength of texture enhancement according to their preferences.

\begin{table}[t]
\centering
\caption{Comparison of alternative training and conditioning strategies on the RealSR dataset. Top: LoRA training strategies for pixel-, semantic-, and texture-level objectives. Bottom: conditioning mechanisms for guiding the diffusion model.}
\vspace{-.6em}
\setlength{\tabcolsep}{0.3em}
\resizebox{\linewidth}{!}{
\begin{tabular}{lcccccc}
\toprule
 & PSNR $\uparrow$ & SSIM $\uparrow$ & LPIPS $\downarrow$ & DISTS $\downarrow$ & FID $\downarrow$ & NIQE $\downarrow$ \\
\midrule
\rowcolor{TitleColor}
\multicolumn{7}{l}{\textit{Training Variants}} \\
Single LoRA (Joint Training) & 26.66 & 76.15 & 23.91 & 18.17 & 103.66 & \textbf{5.75}\\ 
Single LoRA (Merged) & \underline{26.75} & \underline{76.59} & \textbf{22.44} & \textbf{17.47} & \underline{103.25} & 6.16\\ 
Two LoRAs ($\mathcal{L}_\text{gram}$ on Both) & 26.39 & 75.06 & 24.45 & 18.87 & 109.01 & \underline{5.64}\\ 
\rowcolor{OurColor}
Three LoRAs (\textbf{\ours}) & \textbf{26.81} & \textbf{76.68} & \underline{23.05} & \underline{17.91} & \textbf{100.71} & 5.84 \\
\midrule
\rowcolor{TitleColor}
\multicolumn{7}{l}{\textit{Conditioning Variants}} \\
Fixed Conditioning Tensor & \underline{26.37} & \underline{75.54} & 25.19 & \underline{18.90} & \underline{110.56} & \underline{5.53}\\ 
Learnable Conditioning Tensor  & 26.26 & 75.47 & \underline{25.09} & 19.19 & 113.05 & \textbf{5.46}\\ 
\rowcolor{OurColor}
Visual Conditioning (\textbf{\ours}) & \textbf{26.81} & \textbf{76.68} & \textbf{23.05} & \textbf{17.91} & \textbf{100.71} & 5.84 \\
\bottomrule
\end{tabular}
}
\label{tab:training_strategies}
\vspace{-0.35cm}
\end{table}

\tit{Effect of LoRA Training Strategy} Table~\ref{tab:training_strategies} (top) compares alternative strategies for incorporating pixel-, semantic-, and texture-level objectives through LoRA adaptation. Training a single LoRA with all losses jointly achieves competitive PSNR and SSIM but performs worse on perceptual metrics such as DISTS and FID, indicating limited texture consistency. Merging the three LoRAs into a single module and fine-tuning it using the same loss of the $\Delta\theta_{gram}$ LoRA leads to marginal gains on PSNR and SSIM, but does not improve perceptual quality. Applying the Gram loss to both pixel- and semantic-level LoRAs further degrades performance, suggesting that texture alignment interferes with early-stage restoration when not properly isolated. In contrast, the proposed three-LoRA strategy consistently achieves the best results, yielding the lowest LPIPS, DISTS, and FID while maintaining the highest PSNR and SSIM. These results show that sequentially separating degradation removal, semantic enhancement, and texture alignment is crucial for effective optimization.

\tit{Effect of Conditioning Design}
Table~\ref{tab:training_strategies} (bottom) evaluates different conditioning mechanisms used to guide the diffusion model. Replacing semantic conditioning with a fixed conditioning tensor results in a noticeable drop in both fidelity and perceptual metrics. Allowing this tensor to be learnable yields only marginal improvements and remains significantly inferior to the proposed approach. In contrast, visual conditioning with DINOv3 features consistently outperforms parametric conditioning across all metrics, achieving substantial gains in PSNR, SSIM, LPIPS, DISTS, and FID. This confirms that dense visual features extracted from the low-resolution input provide more informative and spatially aligned guidance than simple learned conditioning parameters.

\begin{table}[t]
\centering
\caption{Generalization across different visual encoders on the RealSR dataset. Performance of \ours using various frozen visual backbones for conditioning and Gram matrix alignment, compared to the baseline text-conditioned two-LoRA model. Best results are in bold, second best are underlined.}
\vspace{-.6em}
\setlength{\tabcolsep}{.45em}
\begin{tabular}{lccccccc}
\toprule
 & \textbf{Backbone} & PSNR $\uparrow$ & SSIM $\uparrow$ & LPIPS $\downarrow$ & DISTS $\downarrow$ & FID $\downarrow$ & NIQE $\downarrow$ \\
\midrule
Baseline & - & 25.48 & 74.15 & 26.72 & 20.42 & 124.34 & \textbf{5.51}\\
\midrule
\multirow{6}{*}{\textbf{\ours}} & CLIP-B & 26.67 & 76.28 & 24.06 & 18.66 & 107.35 & 5.93\\ 
& SigLIP2-B & 26.67 & 76.15 & 23.51 & 18.11 & 104.42 & \underline{5.77}\\ 
& DINOv2-B & 26.61 & 75.75 & \underline{23.49} & \textbf{17.78} & 104.49 & 5.97\\ 
& DINOv3-S+  & \underline{26.78} & \underline{76.58} & 23.70 & 18.36 & 104.73 & 5.79\\ 
& DINOv3-B & \textbf{26.81} & \textbf{76.68} & \textbf{23.05} & \underline{17.91} & \underline{100.71} & 5.84 \\
& DINOv3-L & 26.64 & 76.27 & 23.53 & 18.07 & \textbf{99.62} & 5.85\\ 
\bottomrule
\end{tabular}
\label{tab:encoder}
\vspace{-0.35cm}
\end{table}

\subsection{Generalization Across Visual Encoders} 
Table~\ref{tab:encoder} evaluates the generalization ability of \ours when using different frozen visual encoders for conditioning and texture alignment on the RealSR dataset. All variants are compared against a text-conditioned two-LoRA baseline. Overall, replacing text conditioning with visual conditioning consistently improves performance across all metrics. Using visual encoders yields substantial gains in PSNR and SSIM while significantly reducing LPIPS, DISTS, and FID, demonstrating improved structural fidelity and perceptual quality. 

Among the evaluated backbones, multimodal encoders such as CLIP-B~\cite{radford2021learningtransferablevisualmodels} and SigLIP2-B~\cite{tschannen2025siglip2multilingualvisionlanguage} provide clear improvements over the baseline, indicating that high-level semantic visual representations are beneficial for guiding diffusion-based SR. However, encoders designed for dense visual representation learning achieve stronger results. In particular, DINOv2-B~\cite{oquab2024dinov} and the DINOv3~\cite{simeoni2025dinov3} family consistently outperform multimodal alternatives, highlighting the importance of spatially detailed and texture-aware features.

Within the DINOv3 family, the base variant achieves the best overall performance. Notably, increasing model size does not lead to monotonic improvements, as both smaller and larger variants perform slightly worse than DINOv3-B. This suggests that an appropriate balance between representation capacity and spatial precision is more critical than model scale for conditioning diffusion-based SR. These results demonstrate that \ours generalizes well across different visual encoders and that strong performance does not rely on a specific backbone choice, while DINOv3-B provides the most effective trade-off across metrics.
\section{Conclusion}
We presented a diffusion-based SR framework that replaces text conditioning with visual feature conditioning and introduces texture-aware optimization through Gram matrix alignment. By leveraging DINOv3 features and a three-stage LoRA training strategy, the proposed method effectively disentangles degradation removal, semantic enhancement, and texture alignment. Extensive experiments demonstrate consistent improvements over state-of-the-art one-step diffusion methods, particularly on real-world benchmarks. Overall, our results show that explicitly modeling visual structure and texture consistency can significantly improve diffusion-based SR.

\subsubsection{Acknowledgments} This work has been conducted under a research grant co-funded by Maticad s.r.l. and supported by the EU Horizon project ELIAS (No. 101120237). We further acknowledge the CINECA award, under the ISCRA initiative, for the availability of high-performance computing resources.

%
%
%
\bibliographystyle{splncs04}
\bibliography{bibliography}

@string{cvpr     = {CVPR}}

@string{cvprw    = {CVPR Workshops}}

@string{nips     = {NeurIPS}}

@string{iccv     = {ICCV}}

@string{eccv     = {ECCV}}

@string{iclr     = {ICLR}}

@string{icml     = {ICML}}

@string{ieeetpami  = {IEEE Trans. PAMI}}

@string{ieeetip    = {IEEE Trans. Image Processing}}

@string{ijcv     = {IJCV}}

@string{tmlr      = {TMLR}}

@inproceedings{sun2025pixel,
  title={{Pixel-level and Semantic-level Adjustable Super-resolution: A Dual-LoRA Approach}},
  author={Sun, Lingchen and Wu, Rongyuan and Ma, Zhiyuan and Liu, Shuaizheng and Yi, Qiaosi and Zhang, Lei},
  booktitle=cvpr,
  year={2025}
}

@article{simeoni2025dinov3,
  title={{DINOv3}},
  author={Sim{\'e}oni, Oriane and Vo, Huy V and Seitzer, Maximilian and Baldassarre, Federico and Oquab, Maxime and Jose, Cijo and Khalidov, Vasil and Szafraniec, Marc and Yi, Seungeun and Ramamonjisoa, Micha{\"e}l and others},
  journal={arXiv preprint arXiv:2508.10104},
  year={2025}
}

@article{
oquab2024dinov,
title={{DINOv2: Learning Robust Visual Features without Supervision}},
author={Oquab, Maxime and Darcet, Timoth{\'e}e and Moutakanni, Th{\'e}o and Vo, Huy and Szafraniec, Marc and Khalidov, Vasil and Fernandez, Pierre and Haziza, Daniel and Massa, Francisco and El-Nouby, Alaaeldin and others},
journal=tmlr,
year={{2024}}
}

@inproceedings{
yu2024texttod,
title={{Text-to-3D with Classifier Score Distillation}},
author={Yu, Xin and Guo, Yuan-Chen and Li, Yangguang and Liang, Ding and Zhang, Song-Hai and Qi, Xiaojuan},
booktitle=iclr,
year={{2024}},
}

@inproceedings{li2023lsdir,
  title={{LSDIR: A Large Scale Dataset for Image Restoration}},
  author={Li, Yawei and Zhang, Kai and Liang, Jingyun and Cao, Jiezhang and Liu, Ce and Gong, Rui and Zhang, Yulun and Tang, Hao and Liu, Yun and Demandolx, Denis and others},
  booktitle=cvpr,
  year={2023}
}

@inproceedings{bai2023ffhq,
  title={{FFHQ-UV: Normalized Facial UV-Texture Dataset for 3D Face Reconstruction}},
  author={Bai, Haoran and Kang, Di and Zhang, Haoxian and Pan, Jinshan and Bao, Linchao},
  booktitle=cvpr,
  year={2023}
}

@inproceedings{wang2021real,
  title={{Real-ESRGAN: Training Real-World Blind Super-Resolution With Pure Synthetic Data}},
  author={Wang, Xintao and Xie, Liangbin and Dong, Chao and Shan, Ying},
  booktitle=iccv,
  year={2021}
}

@inproceedings{agustsson2017ntire,
  title={{NTIRE 2017 Challenge on Single Image Super-Resolution: Dataset and Study}},
  author={Agustsson, Eirikur and Timofte, Radu},
  booktitle=cvprw,
  year={2017}
}

@inproceedings{cai2019toward,
  title={{Toward Real-World Single Image Super-Resolution: A New Benchmark and a New Model}},
  author={Cai, Jianrui and Zeng, Hui and Yong, Hongwei and Cao, Zisheng and Zhang, Lei},
  booktitle=iccv,
  year={2019}
}

@inproceedings{wei2020component,
  title={{Component Divide-and-Conquer for Real-World Image Super-Resolution}},
  author={Wei, Pengxu and Xie, Ziwei and Lu, Hannan and Zhan, Zongyuan and Ye, Qixiang and Zuo, Wangmeng and Lin, Liang},
  booktitle=eccv,
  year={2020},
}

@article{wang2004image,
  title={Image quality assessment: from error visibility to structural similarity},
  author={Wang, Zhou and Bovik, Alan C and Sheikh, Hamid R and Simoncelli, Eero P},
  journal=ieeetip,
  volume={13},
  number={4},
  pages={600--612},
  year={2004},
}

@inproceedings{zhang2018unreasonable,
  title={{The Unreasonable Effectiveness of Deep Features as a Perceptual Metric}},
  author={Zhang, Richard and Isola, Phillip and Efros, Alexei A and Shechtman, Eli and Wang, Oliver},
  booktitle=cvpr,
  year={2018}
}

@article{ding2020image,
  title={{Image quality assessment: Unifying structure and texture similarity}},
  author={Ding, Keyan and Ma, Kede and Wang, Shiqi and Simoncelli, Eero P},
  volume={44},
  number={5},
  pages={2567--2581},
  journal=ieeetpami,
  year={2020}
}

@inproceedings{heusel2017gans,
  title={{GANs Trained by a Two Time-Scale Update Rule Converge to a Local Nash Equilibrium}},
  author={Heusel, Martin and Ramsauer, Hubert and Unterthiner, Thomas and Nessler, Bernhard and Hochreiter, Sepp},
  booktitle=nips,
  year={2017}
}

@article{mittal2012making,
  title={{Making a ``completely blind'' image quality analyzer}},
  author={Mittal, Anish and Soundararajan, Rajiv and Bovik, Alan C},
  volume={20},
  number={3},
  pages={209--212},
  journal={IEEE Signal Processing Letters},
  year={2012},
}

@inproceedings{wu2024one,
  title={{One-Step Effective Diffusion Network for Real-World Image Super-Resolution}},
  author={Wu, Rongyuan and Sun, Lingchen and Ma, Zhiyuan and Zhang, Lei},
  booktitle=nips,
  year={2024}
}

@inproceedings{dong2014learning,
  title={{Learning a Deep Convolutional Network for Image Super-Resolution}},
  author={Dong, Chao and Loy, Chen Change and He, Kaiming and Tang, Xiaoou},
  booktitle=eccv,
  year={2014},
}

@inproceedings{tong2017image,
  title={{Image Super-Resolution Using Dense Skip Connections}},
  author={Tong, Tong and Li, Gen and Liu, Xiejie and Gao, Qinquan},
  booktitle=iccv,
  year={2017}
}

@inproceedings{lim2017enhanced,
  title={{Enhanced Deep Residual Networks for Single Image Super-Resolution}},
  author={Lim, Bee and Son, Sanghyun and Kim, Heewon and Nah, Seungjun and Mu Lee, Kyoung},
  booktitle=cvprw,
  year={2017}
}

@inproceedings{zhang2021designing,
  title={{Designing a Practical Degradation Model for Deep Blind Image Super-Resolution}},
  author={Zhang, Kai and Liang, Jingyun and Van Gool, Luc and Timofte, Radu},
  booktitle=iccv,
  year={2021}
}

@inproceedings{rombach2022high,
  title={{High-Resolution Image Synthesis With Latent Diffusion Models}},
  author={Rombach, Robin and Blattmann, Andreas and Lorenz, Dominik and Esser, Patrick and Ommer, Bj{\"o}rn},
  booktitle=cvpr,
  year={2022}
}

@article{wang2024exploiting,
  title={{Exploiting Diffusion Prior for Real-World Image Super-Resolution}},
  author={Wang, Jianyi and Yue, Zongsheng and Zhou, Shangchen and Chan, Kelvin CK and Loy, Chen Change},
  volume={132},
  number={12},
  pages={5929--5949},
  journal=ijcv,
  year={2024},
}

@inproceedings{yang2024pixel,
  title={{Pixel-Aware Stable Diffusion for Realistic Image Super-Resolution and Personalized Stylization}},
  author={Yang, Tao and Wu, Rongyuan and Ren, Peiran and Xie, Xuansong and Zhang, Lei},
  booktitle=eccv,
  year={2024},
}

@inproceedings{wang2024sinsr,
  title={{SinSR: Diffusion-Based Image Super-Resolution in a Single Step}},
  author={Wang, Yufei and Yang, Wenhan and Chen, Xinyuan and Wang, Yaohui and Guo, Lanqing and Chau, Lap-Pui and Liu, Ziwei and Qiao, Yu and Kot, Alex C and Wen, Bihan},
  booktitle=cvpr,
  year={2024}
}

@inproceedings{yue2023resshift,
  title={{ResShift: Efficient Diffusion Model for Image Super-resolution by Residual Shifting}},
  author={Yue, Zongsheng and Wang, Jianyi and Loy, Chen Change},
  booktitle=nips,
  year={2023}
}

@inproceedings{johnson2016perceptual,
  title={{Perceptual Losses for Real-Time Style Transfer and Super-Resolution}},
  author={Johnson, Justin and Alahi, Alexandre and Fei-Fei, Li},
  booktitle=eccv,
  year={2016},
}

@inproceedings{ledig2017photo,
  title={{Photo-Realistic Single Image Super-Resolution Using a Generative Adversarial Network}},
  author={Ledig, Christian and Theis, Lucas and Husz{\'a}r, Ferenc and Caballero, Jose and Cunningham, Andrew and Acosta, Alejandro and Aitken, Andrew and Tejani, Alykhan and Totz, Johannes and Wang, Zehan and others},
  booktitle=cvpr,
  year={2017}
}

@inproceedings{liang2022details,
  title={{Details or Artifacts: A Locally Discriminative Learning Approach to Realistic Image Super-Resolution}},
  author={Liang, Jie and Zeng, Hui and Zhang, Lei},
  booktitle=cvpr,
  year={2022}
}

@article{xie2023desra,
  title={{DeSRA: Detect and Delete the Artifacts of GAN-based Real-World Super-Resolution Models}},
  author={Xie, Liangbin and Wang, Xintao and Chen, Xiangyu and Li, Gen and Shan, Ying and Zhou, Jiantao and Dong, Chao},
  journal={arXiv preprint arXiv:2307.02457},
  year={2023}
}

@inproceedings{dhariwal2021diffusion,
  title={{Diffusion Models Beat GANs on Image Synthesis}},
  author={Dhariwal, Prafulla and Nichol, Alexander},
  booktitle=nips,
  year={2021}
}

@inproceedings{song2020score,
  title={{Score-Based Generative Modeling through Stochastic Differential Equations}},
  author={Song, Yang and Sohl-Dickstein, Jascha and Kingma, Diederik P and Kumar, Abhishek and Ermon, Stefano and Poole, Ben},
  booktitle=iclr,
  year={2021}
}

@inproceedings{wu2024seesr,
  title={{SeeSR: Towards Semantics-Aware Real-World Image Super-Resolution}},
  author={Wu, Rongyuan and Yang, Tao and Sun, Lingchen and Zhang, Zhengqiang and Li, Shuai and Zhang, Lei},
  booktitle=cvpr,
  year={2024}
}

@article{sun2023improving,
  title={{Improving the Stability and Efficiency of Diffusion Models for Content Consistent Super-Resolution}},
  author={Sun, Lingchen and Wu, Rongyuan and Liang, Jie and Zhang, Zhengqiang and Yong, Hongwei and Zhang, Lei},
  journal={arXiv preprint arXiv:2401.00877},
  year={2023}
}

@inproceedings{yu2024scaling,
  title={{Scaling Up to Excellence: Practicing Model Scaling for Photo-Realistic Image Restoration In the Wild}},
  author={Yu, Fanghua and Gu, Jinjin and Li, Zheyuan and Hu, Jinfan and Kong, Xiangtao and Wang, Xintao and He, Jingwen and Qiao, Yu and Dong, Chao},
  booktitle=cvpr,
  year={2024}
}

@article{arora2025guidesrrethinkingguidanceonestep,
  title={{GuideSR: Rethinking Guidance for One-Step High-Fidelity Diffusion-Based Super-Resolution}},
  author={Arora, Aditya and Tu, Zhengzhong and Wang, Yufei and Bai, Ruizheng and Wang, Jian and Ma, Sizhuo},
  journal={arXiv preprint arXiv:2505.00687},
  year={2025}
}

@article{chai2025omniscales,
  title={{OmniScaleSR: Unleashing Scale-Controlled Diffusion Prior for Faithful and Realistic Arbitrary-Scale Image Super-Resolution}},
  author={Chai, Xinning and Cheng, Zhengxue and Zhang, Yuhong and Zhang, Hengsheng and Qin, Yingsheng and Yang, Yucai and Xie, Rong and Song, Li},
  journal={arXiv preprint arXiv:2512.04699},
  year={2025}
}

@inproceedings{Chen2024AdversarialDC,
  title={{Adversarial Diffusion Compression for Real-World Image Super-Resolution}},
  author={Chen, Bin and Li, Gehui and Wu, Rongyuan and Zhang, Xindong and Chen, Jie and Zhang, Jian and Zhang, Lei},
  booktitle=cvpr,
  year={2025}
}

@article{Sun2025PocketSRTS,
  title={{PocketSR: The Super-Resolution Expert in Your Pocket Mobiles}},
  author={Sun, Haoze and Jiang, Linfeng and Li, Fan and Pei, Renjing and Wang, Zhixin and Guo, Yong and Xu, Jiaqi and Chen, Haoyu and Han, Jin and Song, Fenglong and others},
  journal={arXiv preprint arXiv:2510.03012},
  year={2025}
}

@article{Wu2025DP2OSRDP,
  title={{DP$^2$O-SR: Direct Perceptual Preference Optimization for Real-World Image Super-Resolution}},
  author={Wu, Rongyuan and Sun, Lingchen and Zhang, Zhengqiang and Wang, Shihao and Wu, Tianhe and Yi, Qiaosi and Li, Shuai and Zhang, Lei},
  journal={arXiv preprint arXiv:2510.18851},
  year={2025}
}

@article{Wang2024SAMDiffSRSD,
  title={{SAM-DiffSR: Structure-Modulated Diffusion Model for Image Super-Resolution}},
  author={Wang, Chengcheng and Hao, Zhiwei and Tang, Yehui and Guo, Jianyuan and Yang, Yujie and Han, Kai and Wang, Yunhe},
  journal={arXiv preprint arXiv:2402.17133},
  year={2024}
}

@article{Li2024BlindDiffED,
  title={{BlindDiff: Empowering Degradation Modelling in Diffusion Models for Blind Image Super-Resolution}},
  author={Li, Feng and Wu, Yixuan and Liang, Zichao and Cong, Runmin and Bai, Huihui and Zhao, Yao and Wang, Meng},
  journal={arXiv preprint arXiv:2403.10211},
  year={2024}
}

@inproceedings{radford2021learningtransferablevisualmodels,
  title={{Learning Transferable Visual Models From Natural Language Supervision}}, 
  author={Radford, Alec and Kim, Jong Wook and Hallacy, Chris and Ramesh, Aditya and Goh, Gabriel and Agarwal, Sandhini and Sastry, Girish and Askell, Amanda and Mishkin, Pamela and Clark, Jack and others},
  booktitle=icml,
  year={2021},
}

@article{tschannen2025siglip2multilingualvisionlanguage,
  title={{SigLIP 2: Multilingual Vision-Language Encoders with Improved Semantic Understanding, Localization, and Dense Features}}, 
  author={Tschannen, Michael and Gritsenko, Alexey and Wang, Xiao and Naeem, Muhammad Ferjad and Alabdulmohsin, Ibrahim and Parthasarathy, Nikhil and Evans, Talfan and Beyer, Lucas and Xia, Ye and Mustafa, Basil and others},
  journal={arXiv preprint arXiv:2502.14786},
  year={2025}
}

@article{zhang2024degradation,
  title={Degradation-guided one-step image super-resolution with diffusion priors},
  author={Zhang, Aiping and Yue, Zongsheng and Pei, Renjing and Ren, Wenqi and Cao, Xiaochun},
  journal={arXiv preprint arXiv:2409.17058},
  year={2024}
}

@inproceedings{hu2022lora,
  title={{LoRA: Low-Rank Adaptation of Large Language Models}},
  author={Hu, Edward J and Shen, Yelong and Wallis, Phillip and Allen-Zhu, Zeyuan and Li, Yuanzhi and Wang, Shean and Wang, Lu and Chen, Weizhu and others},
  booktitle=iclr,
  year={2022}
}

@inproceedings{li2022blip,
  title={{BLIP: Bootstrapping Language-Image Pre-training for Unified Vision-Language Understanding and Generation}},
  author={Li, Junnan and Li, Dongxu and Xiong, Caiming and Hoi, Steven},
  booktitle=icml,
  year={2022},
}

@inproceedings{li2023blip2,
  title={{BLIP-2: Bootstrapping Language-Image Pre-training with Frozen Image Encoders and Large Language Models}},
  author={Li, Junnan and Li, Dongxu and Savarese, Silvio and Hoi, Steven},
  booktitle=icml,
  year={2023},
}

@inproceedings{gatys2016image,
  title={{Image Style Transfer Using Convolutional Neural Networks}},
  author={Gatys, Leon A and Ecker, Alexander S and Bethge, Matthias},
  booktitle=cvpr,
  year={2016}
}
\end{document}